\documentclass[11pt,a4paper]{article}
\usepackage[utf8]{inputenc}
\usepackage[hyperref]{emnlp2018}
\usepackage{times}
\usepackage{latexsym}
\usepackage{tikz}
\usepackage{gnuplot-lua-tikz}
\usepackage{todonotes}
\usepackage{arydshln}
\usepackage{amsmath}
\usepackage{amsfonts}
\usepackage{amssymb}

\usepackage{url}

\aclfinalcopy 
\def\aclpaperid{1490} 

\title{End-to-End Non-Autoregressive Neural Machine Translation 
\\with Connectionist Temporal Classification}

\author{Jind\v rich Libovick\' y \and Jind\v rich Helcl \\
  Charles University, Faculty of Mathematics and Physics \\
  Institute of Formal and Applied Linguistics \\
  Malostransk\' e n\' am\v est\' i 25, 118 00 Prague, Czech Republic \\
  {\tt \{libovicky, helcl\}@ufal.mff.cuni.cz} \\}

\date{}

\begin{document}
\maketitle
\begin{abstract}
Autoregressive decoding is the only part of sequence-to-sequence
models that prevents them from massive parallelization at inference time.
Non-autoregressive models enable the decoder to generate all output
symbols independently in parallel.
We present a novel non-autoregressive architecture based on connectionist
temporal classification and evaluate it on the task of neural machine translation.
Unlike other non-autoregressive methods which operate in several steps,
our model can be trained end-to-end.
We conduct experiments on the WMT English-Romanian and English-German datasets.
Our models achieve a significant speedup over the autoregressive models, keeping the 
translation quality comparable to other non-autoregressive models.
\end{abstract}

\section{Introduction}

Parallelization is the key ingredient for making deep learning models
computationally tractable. While the advantages of parallelization are 
exploited on many levels during training and inference, autoregressive
decoders require sequential execution.

Training and inference algorithms in sequence-to-sequence tasks with recurrent 
neural networks (RNNs) such as neural machine translation (NMT) have
linear time complexity w.r.t. the target sequence length, even when parallelized
\citep{sutskever2014sequence, bahdanau2015neural}.

Recent approaches such as convolutional sequence-to-sequence learning
\citep{gehring2017convolutional} or self-attentive networks a.k.a.~the~Transformer
\citep{vaswani2017attention} replace RNNs with parallelizable components
in order to reduce the time complexity of the training.
In these models, the decoding is still sequential, because
the probability of emitting a symbol is conditioned on the previously decoded
symbols.

In non-autoregressive decoders, the inference algorithm can be parallelized
because the decoder does not depend on its previous outputs. 
The apparent advantage of this approach is the near-constant time complexity achieved by
the parallelization.
On the other hand, the drawback is that the model needs to explicitly determine the 
target sentence length and reorder the state sequence before it starts generating 
the output. In the current research contributions on this topic, these parts are
trained separately and the inference is done in several steps.







In this paper, we propose an end-to-end non-autoregressive model for NMT using 
Connectionist Temporal Classification (CTC; \citealt{graves2006connectionist}).
The proposed technique achieves promising results on translation between 
English-Romanian and English-German on the WMT News task datasets.

The paper is organized as follows. In Section~\ref{sec:related}, we summarize the 
related work on non-autoregressive NMT. Section~\ref{sec:architecture}
describes the architecture of our proposed model. Section~\ref{sec:experiments} 
presents details of the conducted experiments. The results are discussed in Section~\ref{sec:results}.
We conclude and present ideas for future work in Section~\ref{sec:conclusions}.

\section{Non-Autoregressive NMT}
\label{sec:related}

In this section, we describe two methods for non-autoregressive decoding in NMT.
Both of them are based on the Transformer architecture \citep{vaswani2017attention}, 
with the encoder part unchanged.

\citet{gu2017nonautoregressive} use a latent fertility model to copy the 
sequence of source embeddings which is then used for the target sentence 
generation. The fertility (i.e. the number of target words for each source word)
is estimated using a softmax on the encoder states.
In the decoder, the input embeddings are repeated based on their fertility. The~decoder
has the same architecture as the encoder plus the encoder attention.
The best results were achieved by sampling fertilities from the model and then rescoring
the output sentences using an autoregressive model. The reported inference speed of this 
method is 2--15 times faster than of a comparable autoregressive model,
depending on the number of fertility samples.

\citet{lee2018deterministic} propose an architecture with two decoders. The 
first decoder generates a candidate translation from a source sentence padded to an 
estimated target length. The explicit length estimate is done with a softmax
over possible sentence lengths (up to a fixed maximum). The output of the
first decoder is then fed as an input to the second decoder.
The second decoder is used as a denoising auto-encoder and can be
applied iteratively. Both decoders have the same
architecture as in \citet{gu2017nonautoregressive}. They achieved a speedup of 16 times
over the autoregressive model with a single denoising iteration. They report the best result
in terms of BLEU~\citep{papineni2002bleu} after 20 iterations with almost no inference speedup 
compared to their autoregressive baseline.


\section{Proposed Architecture}
\label{sec:architecture}

\begin{figure}
\begin{center}
\scalebox{0.75}{\def\inputsize{7}

\begin{tikzpicture}[]

\draw (\inputsize / 2 + 0.1, -0.1) node {Input token embeddings};

\foreach \i in {0,...,\inputsize} {
	\draw (\i,-0.5) rectangle (\i+0.2,-1);
    \draw [->] (\i+0.1,-1) -- (\i+0.1, -1.25); 
};

\draw (0, -1.25) rectangle (\inputsize + 0.2, -2.25);
\draw (\inputsize / 2 + 0.1, -1.75) node {Encoder};

\foreach \i in {0,...,\inputsize} {
	\draw [->] (\i+0.1,-2.25) -- (\i+0.1, -2.5); 
    \draw[fill=yellow!40] (\i,-2.5) rectangle (\i+0.2,-3);

    \draw [->] (\i+0.1,-3) -- (\i+0.1, -3.25); 
	\draw[fill=blue!40] (\i,-3.25) rectangle (\i+0.2,-3.75);
	\draw[fill=red!40] (\i,-3.75) rectangle (\i+0.2,-4.25);

    \draw [dashed,->] (\i+0.1,-4.25) -  - (\i+0.1, -4.75); 
    \draw [dashed,->] (\i+0.2,-3.5) .. controls (\i + 0.6, -3.65) .. (\i+0.6, -4.75); 

	\draw[fill=red!40] (\i,-4.75) rectangle (\i+0.2,-5.25);
	\draw[fill=blue!40] (\i + 0.5,-4.75) rectangle (\i+0.7,-5.25);

    \draw [->] (\i+0.1,-5.25) - - (\i+0.1, -5.5); 
    \draw [->] (\i+0.6,-5.25) - - (\i+0.6, -5.5);
};

\draw (\inputsize + 1.2, -2.75) node {$\mathbf{h}$}; 
\draw (\inputsize + 1.2, -3.75) node {$W_{\text{spl}}\mathbf{h}$}; 
\draw (\inputsize + 1.2, -5.00) node {$\mathbf{s}$}; 

\draw (0, -5.5) rectangle (\inputsize + 0.7, -6.5);
\draw (\inputsize / 2 + 0.5 + 0.1, -6.0) node {Decoder};

\draw [fill=green!80!black!60] (\inputsize / 2 + 0.4,-7.2) circle [x radius=\inputsize / 2 + 0.4, y radius=0.5];
\draw (\inputsize / 2 + 0.6, -7.2) node {Connectionist Temporal Classification};

\foreach \i in {0,...,\inputsize} {
   \draw [->] (\i+0.1,-7.9) - - (\i+0.1, -8.15); 
   \draw [->] (\i+0.6,-7.9) - - (\i+0.6, -8.15);
}

\draw  (0+0.1,-8.4) node {$w_1$};
\draw  (0+0.6,-8.4) node {$w_2$};
\draw  (1+0.1,-8.4) node {$w_3$};
\draw  (1+0.6,-8.4) node {$\varnothing$};
\draw  (2+0.1,-8.4) node {$w_4$};
\draw  (2+0.6,-8.4) node {$\varnothing$};
\draw  (3+0.1,-8.4) node {$w_5$};
\draw  (3+0.6,-8.4) node {$w_6$};
\draw  (4+0.1,-8.4) node {$\varnothing$};
\draw  (4+0.6,-8.4) node {$\varnothing$};
\draw  (5+0.1,-8.4) node {$\varnothing$};
\draw  (5+0.6,-8.4) node {$w_7$};
\draw  (6+0.1,-8.4) node {$w_8$};
\draw  (6+0.6,-8.4) node {$\varnothing$};
\draw  (7+0.1,-8.4) node {$w_9$};
\draw  (7+0.6,-8.4) node {$\varnothing$};

\draw (\inputsize / 2 + 0.3, -8.95) node {Output tokens / null symbols};

\end{tikzpicture}}
\end{center}
\caption{Scheme of the proposed architecture. The part between the encoder and the decoder is expressed by Equation~\ref{eq:split}.}
\label{fig:architecture}
\end{figure}
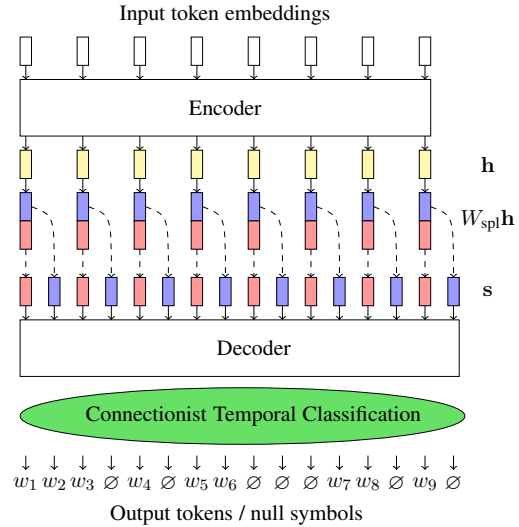

Similar to the previous work \citep{gu2017nonautoregressive,lee2018deterministic},
our models are based on the Transformer architecture as described by 
\citet{vaswani2017attention}, keeping the encoder part unchanged.
Figure~\ref{fig:architecture} illustrates our method and highlights the 
differences from the Transformer model.

In order to generate output words in parallel, we formulate the translation as
a sequence labeling problem. Neural architectures used for encoding input in
NLP tasks usually generate sequences of hidden states of the same or shorter
length as the input sequence. For this reason, we cannot apply the sequence
labeling directly over the states because the target sentence might be longer
than the source sentence.

To enable the labeler to generate sentences that are longer than the source sentence,
we project the encoder output states $\mathbf{h}$ into a $k$-times longer sequence $\mathbf{s}$, such that:
\begin{gather}
s_{ci+b} = \left( W_{\text{spl}} h_{c} + b_{\text{spl}} \right)_{bd:(b+1)d} \label{eq:split}
\end{gather}
for $b = 0 \ldots k-1$, and $c = 0 \ldots T_x$ where $d$ is the Transformer model dimension,
$T_x$ is the length of the source sentence, and $W_{\text{spl}} \in \mathbb{R}^{d \times kd}$ and $b_{\text{spl}} \in \mathbb{R}^{kd}$ are 
trainable projection parameters. In other words, after a linear projection, each state is sliced to $k$ vectors, creating a sequence of length $kT_x$.

In the next step, we process the sequence $\mathbf{s}$ with a decoder. Unlike the
Transformer architecture, our decoder does not use the temporal mask in the 
self-attention step.

Finally, the decoder states are labeled either with an output token or a null symbol.
The number of combinations of the possible positions of the null symbols in the 
output sequence given reference sequence length $T_y$ is $\binom{kT_x}{T_y}$. Because there
is no prior alignment between the input and output symbols, we consider all output sequences
that yield the correct output in the loss function. Because summing the exponential number of combinations directly
is not tractable, we we use the CTC loss \citep{graves2006connectionist} which employs dynamic
programming to compute the negative log-likelihood of the output sequence,
summed over all the combinations. 


The loss can be computed using a linear algorithm similar to training Hidden
Markov Models \citep{rabiner1989tutorial}. The algorithm computes and stores partial log-probabilities sums for all
prefixes and suffixes of the output symbol sequence using dynamic programming.
The table of pre-computed log-probablities allows us to compute the
probability of being a part of a correct output sequence by combining the log-probabilities
of its prefix and suffix.

An appealing property of training using the CTC loss is that the models support
left-to-right beam search decoding by recombining prefixes that yield the same output.
Unlike the greedy decoding this can no longer be done in parallel. 
However, the linear computation is in theory still faster than autoregressive decoding.

\section{Experiments}
\label{sec:experiments}

\begin{table*}
\begin{center}
\scalebox{1.0}{\begin{tabular}{l|cc|cc|cc}
& \multicolumn{2}{c|}{WMT 16} & \multicolumn{2}{c|}{WMT 14} & \multicolumn{2}{c}{WMT 15} \\
& en-ro & ro-en & en-de & de-en & en-de & de-en \\ \hline \hline

autoregressive $b=1$ & 31.93 & 31.55 & 22.71 & 26.39 & 23.40 & 26.49 \\
autoregressive $b=4$ & 32.40 & 32.06 & 23.45 & 27.02 & 24.12 & 27.05 \\ \hline

\citet{gu2017nonautoregressive} greedy
	& 27.29 & 29.06 & 17.69 & 21.47 & --- & ---   \\
\citet{gu2017nonautoregressive} NPD w/ 100 samples 
	& 29.79 & 31.44 & 19.17 & 23.20 & --- & ---   \\
\citet{lee2018deterministic} 1 iteration
    & 24.45 & 23.73 & --- & --- & 12.65 & 14.48 \\
\citet{lee2018deterministic} best result
    & 29.49 & 30.41 & --- & --- & 19.13 & 21.69 \\ \hline \hline
    
our autoregressive $b=1$ & 21.19 & 29.64  & 22.94 & 28.58 & 25.12 & 28.89 \\
\hline

deep encoder 
    & 17.33 & 22.85 & 12.21 & 12.53 & 13.14 & 18.34 \\
\quad + weight averaging 
    & 18.47 & {\bf 24.68} & 14.65 & 16.72 & 16.74 & 18.47 \\
\quad + beam search 
    & 18.70 & 25.28 & 15.19 & 17.58 & 17.59 & 18.70 \\
   \hdashline

encoder-decoder
	& 18.51 & 22.37 & 13.29 & 17.98 & 16.01 & 19.55 \\
\quad + weight averaging 
    & {\bf 19.54} & 24.67 & 16.56 & {\bf 18.64} & 19.46 & {\bf 21.74} \\
\quad + beam search
    & 19.81 & 25.21 & 17.09 & 18.80 & 20.59 & 22.55 \\ 
\hdashline

encoder-decoder w/ pos. encoding 
	& 18.13 & 22.75 & 12.51 & 11.35 & 15.35 & 19.30 \\
\quad + weight averaging 
	& 19.31 & 24.21 & {\bf 17.37} & 18.07 & {\bf 20.30} & 19.64 \\
\quad + beam search
    & 19.93 & 24.71 & 17.68 & 19.80 & 20.67 & 20.43 \\
\end{tabular}}
\end{center}

\caption{Quantitative results in terms of BLEU score of the proposed
methods compared to other non-autoregressive models. Note that our method uses only
a single pass through the network and should be compared with greedy decoding by 
\citet{gu2017nonautoregressive} and 1 model iteration by \citet{lee2018deterministic}.}
\label{tab:results}
\end{table*}

We experiment with three variants of this architecture. All of them have the 
same total number of layers.
First, the \emph{deep encoder} uses a stack of self-attentive layers only.
We apply the state splitting and the labeler on the output of the last encoder layer.
In contrast to Figure~\ref{fig:architecture}, this variant omits the decoder part.
Second, the \emph{encoder-decoder} consists of two stacks of self-attentive layers -- encoder and decoder.
The outputs of the encoder are transformed using Equation~\ref{eq:split}
and processed by the decoder. In each layer, the decoder part attends to the encoder output.
Third, we extend the encoder-decoder variant with positional encoding \citep{vaswani2017attention}.
The positional encoding vectors are added to the decoder input $\mathbf{s}$.

In all the experiments, we used the same hyper-parameters. 
We set the model dimension to 512 and the feed-forward layer dimension to 4096.
We use multi-head attention with 16 heads. 
In the deep encoder setup, we use 12 layers in the encoder, 
in the encoder-decoder setup, we use 6 layers for the encoder and 6 layers for the decoder.
We set the split factor $k$ to 3, so the encoder states are projected to vectors of 1536 units.

We conduct our experiments on English-Romanian and English-German translation. 
These language pairs were selected by the authors of the previous work because
the training datasets for these language pairs are of considerably different sizes.
We follow these choices in order to present comparable results.

For English-Romanian experiments, we used the WMT16 \citep{bojar2016findings} news dataset.
The training data consists of 613k sentence pairs, validation 2k and test 2k.
We used a shared vocabulary of 38k wordpieces \citep{wu2016google,johnson2017google}.

The English-German dataset consists of 4.6M training sentence pairs
from WMT competitions. As a validation set, we used the test set from 
WMT13 \citep{bojar2013findings}, which contains 3k sentence pairs.
To enable comparison to other non-autoregressive approaches, we evaluate our models on 
the test sets from WMT14 \citep{bojar2014findings} with 3k sentence 
pairs and WMT15 \citep{bojar2015findings} with 2.1k sentence pairs. 
As in the previous case, we used shared vocabulary for both languages which contained 41k wordpieces.

The experiments were conducted using Neural Monkey\footnote{\url{https://github.com/ufal/neuralmonkey}} \citep{neuralMonkey}.
We evaluate the models using BLEU score \citep{papineni2002bleu} as implemented in 
SacreBLEU,\footnote{\url{https://github.com/mjpost/sacreBLEU}}
originally a part of the Sockeye toolkit \citep{Sockeye:17}.

\section{Results}
\label{sec:results}

Quantitative results are tabulated in Table~\ref{tab:results}. In general,
our models achieve a similar performance to other non-autoregressive models.
In case of English-German, our results in both directions
are comparable on the WMT~14 test set and slightly better on the WMT~15 test set. This might be given by the fact that our
autoregressive baseline performs better for this language pair than 
for English-Romanian.

The encoder-decoder setup outperforms the deep encoder setup. Including positional encoding
seems beneficial when translating into German. Weight averaging
from the 5 models with the highest validation score during the training
improves the performance consistently.


We performed a manual evaluation on 100 randomly sampled sentences from the
English-German test sets in both directions. The results of the analysis are
summarized in Table~\ref{tab:manual}.

Non-autoregressive translations of sentences that had errors in the
autoregressive translation were often incomprehensible. In general, less than a
quarter of the sentences was completely correct and over two thirds (one half
in the de$\rightarrow$en direction) were comprehensible.  The most frequent
errors include omitting verbs at the end of German sentences and corruption of
named entities and infrequent words that are represented by more wordpieces.
Most of these errors can be attributed to insufficient language-modeling
capabilities of the model.  The results suggest that integrating an external
language model into an efficient beam search implementation could boost the
translation quality while preserving the speedup over the auto-regressive
models.

\begin{table}
\begin{center}
\scalebox{0.9}{\begin{tabular}{l|cc|cc}
		& \multicolumn{2}{c|}{en $\rightarrow$ de} & \multicolumn{2}{c}{de $\rightarrow$ en} \\
        & ~AR~ & NAR & ~AR~ & NAR \\ \hline
Correct        & 65 & 23 & 67 & 13 \\
Comprehensible & 93 & 71 & 92 & 51 \\ \hline
Too short      & 1 & 16 & 0 & 36 \\
Missing verb   & 4 & 35 & 0 & 8 \\
Corrupt. named entity   & 1 & 27 & 8 & 21 \\
Corrupt. other words & 1 & 20 & 0 & 46
\end{tabular}}
\end{center}

\caption{Results of manual evaluation of the autoregressive (AR) and
non-autoregressive (NAR) models (in percents).} 

\label{tab:manual}
\end{table}

\begin{figure}
\scalebox{0.9}{\input{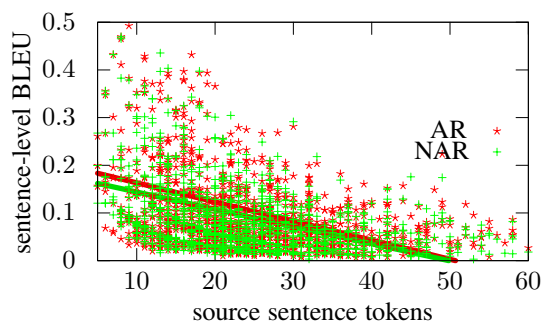}}
    \caption{Comparison of the sentence-level BLEU of our English-to-German
    autoregresssive (AR) and non-autoregressive (NAR) models given the length of the source sentence.}
\label{fig:sentencebleu}
\end{figure}

We also evaluated the translations using sentence-level BLEU score \citep{chen2014sentence}
and measure the Pearson correlation with the length of the source sentence
and the number of null symbols generated in the output. With a growing sentence length,
the scores degrade more in the non-autoregressive model ($r=-0.42$) than in its
autoregressive counterpart ($r=-0.39$). 
The relation between sentence-level BLEU and the source length is plotted in 
Figure~\ref{fig:sentencebleu}.
The sentence-level score is mildly correlated with the number of null symbols in the
non-autoregressive output ($r=0.15$). This suggests that increasing the splitting factor $k$ in 
Equation~\ref{eq:split} might improve the model performance. However, it also 
reduces the efficiency in terms of GPU memory usage.

\begin{figure}
\scalebox{0.9}{\input{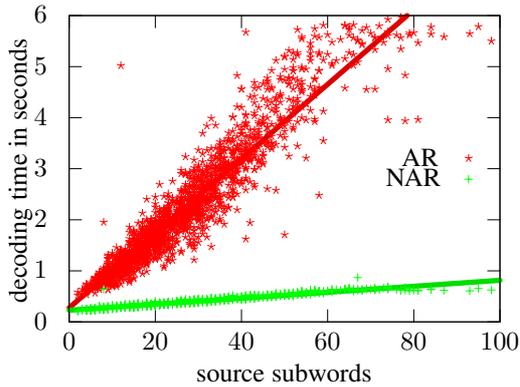}}
\caption{Comparison of CPU decoding time by our autoregressive (AR)
and non-autoregressive (NAR) models based on the source sentence length.}
\label{fig:decodingspeed}
\end{figure}

Figure~\ref{fig:decodingspeed} shows the comparison of the decoding time by
autoregressive and non-autoregressive models. The average times of decoding a
single sentence are shown in Table~\ref{tab:times}. We suspect that the small
difference between CPU and GPU times in the non-autoregressive setup is caused
by the CPU-only implementation of the CTC decoder in TensorFlow
\citep{tensorflow2015-whitepaper}.

\begin{table}
\begin{center}
\scalebox{0.9}{\begin{tabular}{l|c|c}
                & CPU & GPU \\ \hline
 AR, $b=1$      & 2247 ms & 1200 ms  \\
 NAR            & 386 ms  &  350 ms\\
\end{tabular}}
\end{center}

\caption{Average per sentence decoding time for en-de translation.}
\label{tab:times}

\end{table}


\section{Conclusions}
\label{sec:conclusions}

In this work, we presented a novel method for training a non-autoregressive
model end-to-end using connectionist temporal classification. We evaluated the
proposed method on neural machine translation in two language pairs and
compared the results to the previous work.


In general, the results match the translation quality of equivalent variants of the models presented in the previous work. The BLEU score is usually around 80--90\% of the score
of the autoregressive baselines. We measured a 4-times speedup compared to our autoregressive
baseline, which is a smaller gain than reported by the authors of the previous work.
We suspect this might be due to a larger overhead with data loading and processing in Neural Monkey
compared to Tensor2Tensor \citep{tensor2tensor} used by others.

As a future work, we can try to improve the performance of the model by iterative
denoising as done by \citet{lee2018deterministic} while keeping the non-autoregressive 
nature of the decoder.

Another direction of improving the model might be efficient implementation of beam search
which can contain rescoring using an external language model as often done in speech 
recognition \citep{graves2013hybrid}. The non-autoregressive model would play 
a role a of the translation model in the traditional statistical MT problem decomposition.

\section*{Acknowledgments}


This research has been funded by the Czech Science Foundation grant no.
P103/12/G084, Charles
University grant no. 976518 and SVV project no. 260 453.

\bibliography{emnlp2018}
\bibliographystyle{acl_natbib_nourl}

\appendix

\end{document}